# Hybrid A*-Based Reverse Path-Planning of a Vehicle with Trailer System

Xincheng Cao, Haochong Chen, Bilin Aksun-Guvenc, Levent Guvenc
Brian Link, Peter J Richmond, Dokyung Yim, Shihong Fan, John Harber

*Abstract*— Reverse parking maneuvering of a vehicle with trailer system is a difficult task to complete for human drivers due to the multi-body nature of the system and the unintuitive controls required to orientate the trailer properly. The problem is complicated with the presence of other vehicles that the trailer and its connected vehicle must avoid during the reverse parking maneuver. While path planning methods in reverse motion for vehicles with trailers exist, there is a lack of results that also offer collision avoidance as part of the algorithm. This paper hence proposes a modified Hybrid A*-based algorithm that can accommodate the vehicle-trailer system as well as collision avoidance considerations with the other vehicles and obstacles in the parking environment. One of the novelties of this proposed approach is its adaptability to the vehicle with trailer system, where limits of usable steering input that prevent the occurrence of jackknife incidents vary with respect to system configuration. The other contribution is the addition of the collision avoidance functionality which the standard Hybrid A* algorithm lacks. The method is developed and presented first, followed by simulation case studies to demonstrate the efficacy of the proposed approach.

## I. INTRODUCTION

There have been many advancements in the field of autonomous driving vehicles in recent years [1], [2], [3]. A basic functionality of an autonomous vehicle is its capability to plan and track its own path [4], [5]. Reverse parking, particularly in restrictive spaces, remains one of the most challenging maneuvers to complete for an autonomous vehicle and such difficulty increases considerably when the autonomous vehicle is coupled to a trailer unit during this process. A vehicle with a trailer attached needs to be steered in the opposite direction of intended trailer orientation which can be intuitively difficult for inexperienced drivers, requiring the development and utilization of advisory systems to aid drivers [6]. Another challenge is the fact that different steering inputs at the vehicle will be required to orientate the trailer the same way depending on the current pose of the vehicle-trailer combination. Things get more complicated due to the presence of other vehicles that the trailer and its connected vehicle must avoid during the reverse parking maneuver. While path planning methods in reverse motion for vehicles with trailers exist, there is a lack of results that also offer collision avoidance as part of the algorithm which is the motivation underlying the research reported in this paper.

Two types of models have typically been used for vehicle with trailer systems: dynamic models that consider the forces involved and kinematic models that focus on only geometric relationships.

While dynamic models provide more accurate descriptions of the tractor-trailer behaviors, they are typically more complicated in construction. As a result, many studies opt to derive the simpler kinematic model instead, particularly when the maneuvers being studied are low speed in nature, where significant tire deformations do not typically occur [7-10]. The concept of 'virtual tractor', where the last trailer unit in the vehicle-trailer system chain is regarded as a 'virtual tractor' unit has been used in [6], [8], [9], [11], [12].

A collision-free feasible path is the prerequisite of the effective completion of parking maneuvers. Some approaches opt to plan the path that starts from the terminal/goal states and ends at the starting states [13]. Similarly, [14] uses a tree-based path planner while [15] presents a Pontryagin's Minimum Principle (PMP) based optimal path-planning routine with obstacle avoidance through the artificial potential field for forward motion. [16] presents a lattice-based path planner that uses kinematically feasible motion primitives. [17] uses differential flatness and also uses an RRT-based motion planner. [18] explores the minimum parking space for a vehicle with a one-axle trailer. [19] presents feasible paths for the tractor-trailer system by assembling simple path constructs such as rotations, translations, stretches and bends. [20] applies semi-supervised where a deep neural network is used to generate paths that minimize off-track portions of the area swept by the tractor-trailer system. [21] proposes a cooperative trajectory planning algorithm for tractor-trailer wheeled robots.

The above-mentioned path-planning approaches, while effective, are not sufficiently simple in formulation and do not always contain collision avoidance considerations. This paper, thus, proposes a Hybrid A*-based path-planning algorithm that is simple in construction while also providing kinematically feasible and collision-free paths for the automation of vehicle-trailer system reverse parking maneuvers. The outline of the rest of this paper is as follows.

This work was supported in part by Hyundai America Technical Center, Inc. (HATCI). *(Corresponding author: Xincheng Cao).*

Xincheng Cao and Haochong Chen and Bilin Aksun-Guvenc and Levent Guvenc are with the Automated Driving Lab, Department of Mechanical and Aerospace Engineering, The Ohio State University, Columbus, OH 43210, USA. (e-mail: cao.1375@osu.edu; chen.9286@osu.edu; aksunguvenc.1@osu.edu; guvenc.1@osu.edu;).

Brian Link, Peter J Richmond, Dokyung Yim, Shihong Fan and John Harber are with Hyundai America Technical Center, Inc. (HATCI), Superior Township, MI 48198, USA. (e-mail: blink@hatci.com; prichmond@hatci.com; dkyim81@hyundai.com; sfan@hatci.com; jharber@hatci.com).

Section 2 presents the kinematic vehicle-trailer model. Section 3 illustrates the inverse kinematics of the vehicle-trailer system. Section 4 explains the Hybrid A*-based reverse parking path-planning design. Section 5 subsequently presents simulation case study results. The paper ends with conclusions.

## II. KINEMATIC VEHICLE-TRAILER MODEL

Given the low-speed nature of the vehicle-trailer system during reverse parking motion, a kinematic model is used here due to its simplicity and sufficient accuracy. This section aims to derive such a kinematic vehicle-trailer model. The schematic of a generic vehicle-trailer system with one-trailer configuration is shown in Figure 1, and its parameters are listed in Table I. $L_H$, which is the distance between the vehicle rear axle center and the hitch point, determines the type of tractor vehicle (semi-tractor or passenger vehicle), and is a positive value here corresponding to a typical vehicle-trailer system where the vehicle is a passenger car, SUV or pickup truck with a hitch connected to the rear bumper of the vehicle. A front wheel steering vehicle is considered as shown in Figure 1.

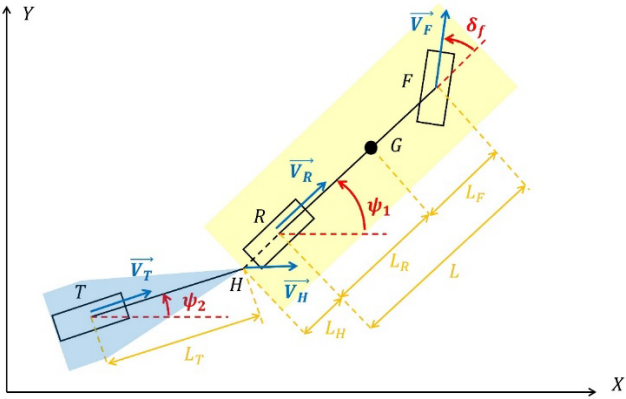

**Fig. 1.** Kinematic vehicle-trailer model with one trailer.

TABLE I
PARAMETERS OF KINEMATIC VEHICLE-TRAILER MODEL

| Model Parameter | Explanation |
|---|---|
| $L$ | Wheelbase of the tractor vehicle (passenger car, SUV or pickup truck) |
| $L_F$ | Distance between vehicle center of gravity G and front axle center |
| $L_R$ | Distance between vehicle center of gravity G and rear axle center |
| $L_H$ | Distance between vehicle rear axle center and trailer hitch joint |
| $L_T$ | Distance between trailer axle center and trailer hitch joint |
| $\delta_f$ | Vehicle front wheel steer angle |
| $\psi_1$ | Vehicle yaw angle |
| $\psi_2$ | Trailer yaw angle |
| $\vec{V_F}$ | Vehicle front axle center velocity |
| $\vec{V_R}$ | Vehicle rear axle center velocity |
| $\vec{V_H}$ | Trailer hitch velocity |
| $\vec{V_T}$ | Trailer axle center velocity |

The kinematic model is
$$\dot{X}_R = V_R \cdot \cos(\psi_1) \quad (1)$$
$$\dot{Y}_R = V_R \cdot \sin(\psi_1) \quad (2)$$
$$\dot{\psi}_1 = \frac{V_R}{L}\tan(\delta_f) \quad (3)$$
$$\dot{X}_T = V_R\cos(\psi_2)[\cos(\Delta\psi) + \frac{L_H}{L}\sin(\Delta\psi)\tan(\delta_f)] \quad (4)$$
$$\dot{Y}_T = V_R\sin(\psi_2)[\cos(\Delta\psi) + \frac{L_H}{L}\sin(\Delta\psi)\tan(\delta_f)] \quad (5)$$
$$\dot{\psi}_2 = \frac{V_R}{L_T}[\sin(\Delta\psi) - \frac{L_H}{L}\cos(\Delta\psi)\tan(\delta_f)] \quad (6)$$

and its derivation is given in references [6]. In this model, the inputs are the vehicle front axle steer angle $\delta_f$ and vehicle rear axle center speed $V_R$.

## III. INVERSE KINEMATICS

### A. Inverse Kinematics Derivation

One of the main challenges of vehicle-trailer reverse parking maneuvers comes from the fact that the system tends to demonstrate unintuitive yaw behavior for inexperienced drivers. As a result, it would be helpful to decouple this multi-body system into individual components, where the trailer unit is treated as a standalone vehicle. One can hence calculate a 'virtual' steering angle at the trailer that would orientate it properly and then map this 'virtual' angle to the actual steer angle at the vehicle steerable axle through kinematic derivation.

The inverse kinematic vehicle-trailer model under consideration is identical to the model illustrated in Figure 1 except for the addition of a 'virtual' steerable axle at the trailer hitch, as demonstrated in Figure 2 that shows only the trailer portion of the model to reduce visual clutter. With this 'virtual' steerable axle, the trailer unit can be regarded as a standalone vehicle, and its 'virtual' steer angle is denoted as $\delta_T$.

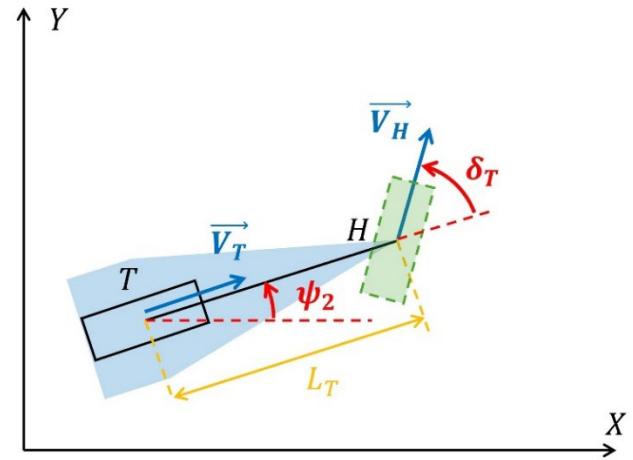

**Fig. 2.** Inverse kinematic vehicle-trailer model: trailer part.

The inverse kinematic model is given by

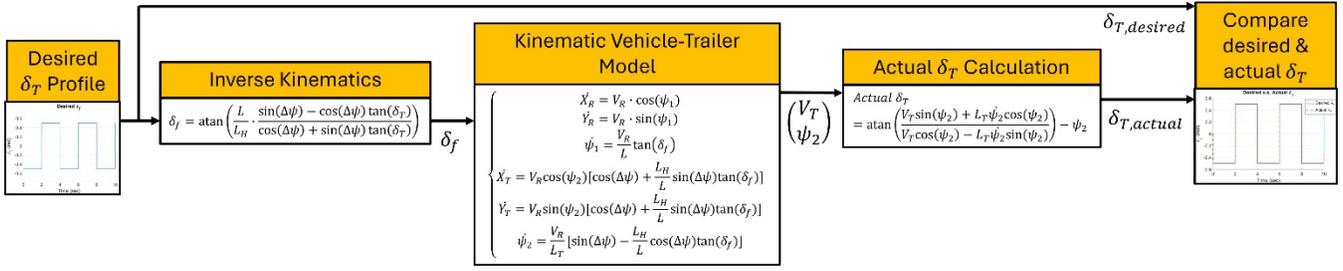

**Fig. 3.** Inverse kinematics simulation study model structure.

$$\dot{\psi}_2 = \frac{V_T}{L_T}\tan(\delta_T) \quad (7)$$

$$\dot{\psi}_1 = \frac{V_T}{L_H}[\sin(\Delta\psi) - \cos(\Delta\psi)\tan(\delta_T)] \quad (8)$$

$$V_T = \frac{V_R}{\cos(\Delta\psi) + \sin(\Delta\psi)\tan(\delta_T)} \quad (9)$$

$$\delta_f = \operatorname{atan}\left(\frac{L}{L_H} \cdot \frac{\sin(\Delta\psi) - \cos(\Delta\psi)\tan(\delta_T)}{\cos(\Delta\psi) + \sin(\Delta\psi)\tan(\delta_T)}\right) \quad (10)$$

where Equations (7) and (8) represent the 'desired' yaw rates of the trailer and the vehicle, respectively, given a 'virtual' steer angle $\delta_T$ while Equation (9) represents the mapping from vehicle rear axle center speed to trailer axle center speed given a 'virtual' steer angle $\delta_T$. Finally, Equation (10) is the mapping equation from the 'virtual' steer angle at the trailer hitch to the actual steer angle at the vehicle steerable (front) axle. The derivation of this inverse kinematic model is given in references [6].

### B. Inverse Kinematics Validation

A simulation study is carried out to demonstrate the effects of the actual-virtual steering angle mapping equation derived in the previous sub-section. The simulation procedure is illustrated in Figure 3. A 'desired' $\delta_T$ profile is first generated and fed into the inverse kinematics calculation block that invokes the actual-virtual steering angle mapping equation, and the resulting vehicle steer angle required is plugged into the kinematic vehicle-trailer model. An additional block is also included to calculate the 'actual' $\delta_T$ based on the outputs of the kinematic vehicle-trailer model, which is then compared to the 'desired' $\delta_T$ profile. Equation (11) illustrates how this 'actual' $\delta_T$ can be calculated.

$$\text{Actual } \delta_T = \operatorname{atan}\left(\frac{Y_H}{X_H}\right) - \psi_2 =$$

$$\operatorname{atan}\left(\frac{V_T\sin(\psi_2) + L_T\dot{\psi}_2\cos(\psi_2)}{V_T\cos(\psi_2) - L_T\dot{\psi}_2\sin(\psi_2)}\right) - \psi_2 \quad (11)$$

The parameter value choices used in this simulation study are detailed in Table II. It is worth noting that this simulation study focuses on the case of reverse motion with the trailer hitch located behind the vehicle rear axle, as this paper limits its scope on vehicle-trailer system backup maneuvers with a car-like (passenger car, SUV or pickup truck) tractor unit. Figure 4 displays the result of the simulation. It can be observed that the vehicle steering inputs generated by the inverse kinematics calculation can accurately re-create the 'desired' $\delta_T$ profile.

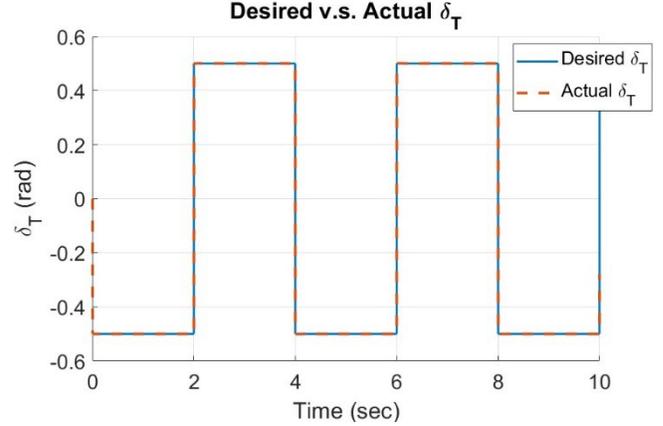

**Fig. 4.** Desired virtual steer angle tracking performance for trailer hitch behind vehicle rear axle in reverse motion.

TABLE II
PARAMETER VALUE CHOICES FOR INVERSE KINEMATICS SIMULATION STUDY

| Model Parameter | Value Choice |
|---|---|
| $L$ | 2.896 [m] |
| $L_H$ | 1.159 [m] (passenger vehicle) |
| $L_T$ | 2.693 [m] |
| $V_R$ | -1 [m/s] (backward motion) |

IV. MULTI-BODY HYBRID A*-BASED PATH-PLANNING DESIGN

### A. Modified Hybrid A* Algorithm Overview

A typical scenario of vehicle-trailer system reverse path-planning operation is shown in Figure 5, where the planning goal is to position the vehicle-trailer system into the desired parking space while avoiding all obstacles that are usually located in neighboring parking spaces. The satisfactory completion of this planning procedure requires both of the following criteria to be met: 1) the generated path must be kinematically feasible for the vehicle with trailer system in reverse motion; 2) the generated path must be collision-free. This is a reverse motion path planning system for a coupled multi-body systems consisting of two bodies, the vehicle and the trailer, with the connection hinge being part of both bodies forming the constraint. To this end, the Hybrid A* algorithm can be used as a beginning of the path-planning approach and extended into a constrained multi-body system with built in collision avoidance. This approach is used as it takes the system kinematics into account. This sub-section hence provides a general overview of a modified Hybrid A* algorithm while later sub-sections dive into design details regarding the multi-body vehicle-trailer system accommodations and collision avoidance considerations.

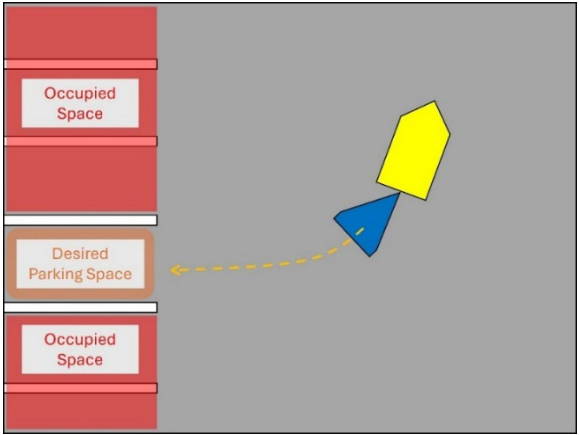

Fig. 5. Typical scenario of vehicle-trailer system reverse parking.

The flowchart of the modified Hybrid A* algorithm is provided in Figure 6. The algorithm works in an iterative manner where several kinematically feasible path branches, called motion primitives, are constructed during each iteration by simulating the kinematic system for a set duration. Assuming that several partially completed paths have already been generated in the previous iterations, the newly generated motion primitives will spawn from the node located at the end of the existing partial path that is closest to the goal. The criteria to determine how close a particular partial path is to the goal is encoded in a metric defined as the cost, where a lower cost value corresponds to an increased proximity to the goal. Once the motion primitive branches are constructed, they are checked for collision, and the end points of the collision-free branches are assigned as the new terminal nodes of their respective partial paths. The terminal node of the partial path with the lowest cost value is subsequently used as the spawning point of additional motion primitive branches in the next iteration, and the cycle continues until a partial branch gets sufficiently close to the goal, in which case this path is regarded as the completed solution path that is both kinematically feasible and also collision-free.

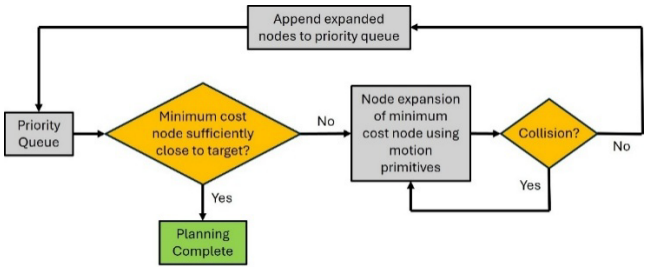

Fig. 6. Modified Hybrid A* algorithm flowchart.

Details of the partial paths, including system states at the terminal nodes and their corresponding cost values, are stored in a priority queue. This queue is ordered by cost values so that the node with the lowest cost can easily be located for motion primitive branch expansion. It should be noted that once the new admissible nodes are added to the queue at the end of each algorithm iteration, the entry used to spawn these new nodes is removed from the queue, as their information will already have been encoded into the new node entries. It is also worth noting that if the lowest cost node entry in the queue cannot yield any collision-free branches, the node entry corresponding to the second lowest cost will be attempted for node expansion operation.

### B. Cost Function Design

As indicated in the algorithm overview, cost values are assigned to the end nodes of partial paths to evaluate their proximity to the goal state. This sub-section aims to explain the design of the cost function in detail.

In general, the cost function in this proposed path-planning algorithm has two components: a heuristic cost element and an accumulated action cost element. The heuristic cost describes how close the trailer state at the partial path end node is to the desired trailer goal state. It is of quadratic form and is displayed in Equation (12).

$$J_H = \left( \begin{bmatrix} X_T \\ Y_T \\ \psi_2 \end{bmatrix} - \begin{bmatrix} X_{T,goal} \\ Y_{T,goal} \\ \psi_{2,goal} \end{bmatrix} \right)^T Q \left( \begin{bmatrix} X_T \\ Y_T \\ \psi_2 \end{bmatrix} - \begin{bmatrix} X_{T,goal} \\ Y_{T,goal} \\ \psi_{2,goal} \end{bmatrix} \right), \text{ where } Q \in \mathbb{R}^{3 \times 3} \succ 0 \quad (12)$$

Apart from the heuristic cost, additional considerations should be given to the selection of the partial path to be expanded. The necessity of this step is demonstrated in Figure 7 where path 1 reaches a query node while path 2 passes through the query node and completes a complete loop before returning to the query node. In this case, path 1 and path 2 have the same heuristic cost values since they both end up at the same terminal states. However, it is more efficient to choose path 1 for further path expansion since it has used less prior expansion steps. As a result, an additional cost function component, referred to as the accumulated action cost $J_A$, is defined in Equation (13), where $N_A$ refers to the number of prior actions already taken and $K_A$ is a scaling factor. Combining this accumulated action cost and the heuristic cost, the overall cost function can be written as Equation (14). It should be remarked that the scaling factor $K_A$ in Equation (13) must be adjusted such that the value of the accumulated action cost is much smaller than that of the heuristic cost since the presence of $J_A$ in the cost function is mainly to differentiate partial paths that have the same terminal states and should not interfere with the metrics of the heuristic cost.

$$J_A = K_A \cdot N_A \quad (13)$$
$$J = J_H + J_A \quad (14)$$

### C. Motion Primitives

As mentioned in a previous sub-section, kinematically feasible motion primitives are generated during each algorithm iteration where several branches spawning from the same node will be constructed. Since this paper focuses on the vehicle with trailer system, the branches are obtained by simulating the kinematic vehicle-trailer system for a set duration. While the number of branches can be custom-defined, this paper presents a three-branch design. The three branches correspond to the following behaviors respectively: 1) backward motion with maximum trailer orientation to the left; 2) backward motion with maximum trailer orientation to the right; 3) backward motion with intermediate trailer orientation (between maximum left and maximum right). All three branches feature the same trailer unit reverse speed $V_T$, which coupled with the same simulation duration, ensures the terminal nodes of these branches being comparable to each other using the cost function mentioned in the previous sub-section. It must also be remarked that each motion primitive branch features constant vehicle-trailer system steering input, the calculation of which is covered in this sub-section as follows.

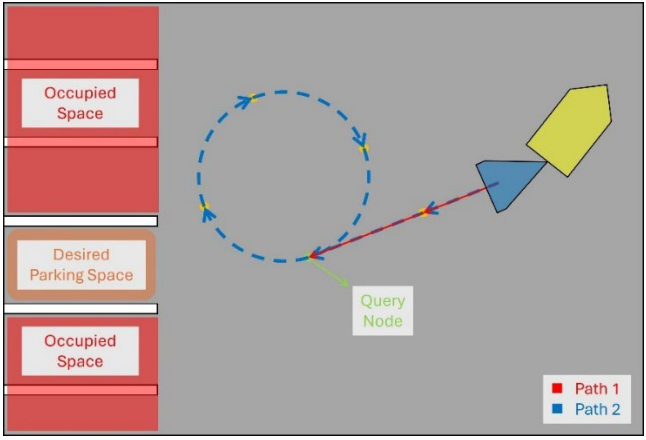

**Fig. 7.** Example of cost function design.

The trailer orientations in maximum left and maximum right directions are determined by the virtual steering limits, denoted as ($\delta_{T,max}$ and $\delta_{T,min}$), which should be carefully designed such that its virtual-actual steering angle mapping under the current hitch angle will not yield an actual steering angle that is unfeasible for the vehicle front axle. At the same time, the virtual steering angle will require its own feasible value range so that it can ensure reasonable trailer orientation behaviors. If we denote the vehicle front wheel steering angle range as $[\delta_{f,min}, \delta_{f,max}]$, then Equation (15) as shown below that maps $\delta_f$ to $\delta_T$ can be used to generate the mapped upper and lower bounds of the virtual steering angle, denoted as $[\delta_{T,lb1}, \delta_{T,ub1}]$, under the current hitch angle.

$$\delta_T = \operatorname{atan}\left(\frac{L \cdot \sin(\Delta\psi) - L_H \cdot \cos(\Delta\psi)\tan(\delta_f)}{L \cdot \cos(\Delta\psi) + L_H \cdot \sin(\Delta\psi)\tan(\delta_f)}\right) \quad (15)$$

If we further define the upper and lower bounds of the virtual steering angle that guarantee reasonable trailer orientation behaviors as $[\delta_{T,lb2}, \delta_{T,ub2}]$, then the virtual steering input constraints ($\delta_{T,min}$ and $\delta_{T,max}$) can be defined as shown in Equation (16).

$$[\delta_{T,min}, \delta_{T,max}] = [\delta_{T,lb1}, \delta_{T,ub1}] \cap [\delta_{T,lb2}, \delta_{T,ub2}] \quad (16)$$

Once the maximum and minimum virtual steering inputs have been obtained, an intermediate virtual steering input, denoted as $\delta_{T,int}$ can be calculated with Equation (17) shown below to generate the motion primitive branch that features the intermediate trailer orientation.

$$\delta_{T,int} = \frac{1}{2}(\delta_{T,max} + \delta_{T,min}) \quad (17)$$

Finally, since the motion primitive branches are generated by simulating the kinematic vehicle-trailer system that takes actual inputs at the vehicle, the above-mentioned virtual inputs at the trailer ($V_T, \delta_T$) must be mapped to the actual inputs at the vehicle ($V_R, \delta_f$) by invoking Equation (9) and Equation (10).

A simple numeric example is presented here to demonstrate the proposed motion primitive design. Model parameter values listed in Table II are reused. If one defines the current hitch angle of the vehicle-trailer system ($\Delta\psi$) and the vehicle front wheel steering angle range ($\delta_{f,min}$ and $\delta_{f,max}$) to take the values listed in Table 3, then applying Equation (15) will yield $[\delta_{T,lb1}, \delta_{T,ub1}] = [-10.447, 30.447]\,[deg]$. If one continues to define a reasonable set of trailer virtual steering angle limits ($\delta_{T,lb2}$ and $\delta_{T,ub2}$) as shown in Table III, one can apply Equation (16) to narrow down the range of admissible virtual steering to $[\delta_{T,min}, \delta_{T,max}] = [-10.447, 28.6479]\,[deg]$. Given this result, the intermediate virtual steering can be calculated as $\delta_{T,int} = 9.1004\,[deg]$ with Equation (17). Further application of Equation (9) and Equation (10) will result in the kinematic vehicle-trailer system inputs that can be used to generate the motion primitive branches.

TABLE III
PARAMETER VALUE CHOICES FOR INPUT CONSTRAINTS DESIGN NUMERIC EXAMPLE

| Model Parameter | Value Choice |
|---|---|
| $\Delta\psi$ | 0.1745 [rad] = 10 [deg] |
| $[\delta_{f,min}, \delta_{f,max}]$ | [-0.75, 0.75] [rad] = [-42.9718, 42.9718] [deg] |
| $[\delta_{T,lb2}, \delta_{T,ub2}]$ | [-0.5, 0.5] [rad] = [-28.6479, 28.6479] [deg] |

*D. Collision Check*

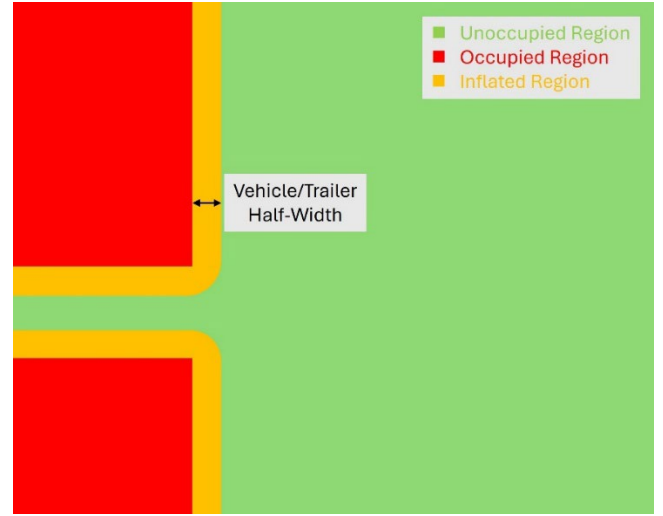

**Fig. 8.** Example of a binary occupancy map.

Once the motion primitives are generated, collision checks will be applied to each candidate branch to determine if these kinematically feasible segments can avoid collisions with obstacles in the parking environment. To facilitate this function, a binary occupancy map should first be constructed to define the regions occupied by obstacles as well as unoccupied regions accessible to the the vehicle-trailer system. It should be remarked that the motion primitive branches generated in the node expansion process represent the trajectories of the vehicle rear axle center and the trailer axle center. This means that collisions may still occur for the outer boundaries of the vehicle-trailer system even if the motion primitives themselves clear the obstacles. Given this consideration, the occupied regions in the binary occupancy map are inflated by the half width of the vehicle or the trailer, whichever is larger, and the inflated region is also assigned as occupied. Figure 8 provides an example of a binary occupancy map with inflated regions illustrated. Given this setup, as long as the vehicle-trailer system centerlines stay within the green unoccupied region, collision avoidance can be guaranteed for both the vehicle unit and the trailer unit.

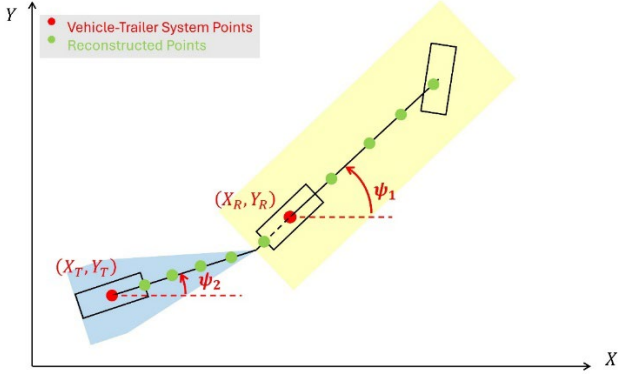

**Fig. 9.** Vehicle-trailer system body reconstruction.

Once the binary occupancy map has been established, the next step is to reconstruct the vehicle-trailer system centerlines. Since the motion primitives already contain vehicle states $(X_R, Y_R, \psi_1)$ and trailer states $(X_T, Y_T, \psi_2)$, the coordinates of additional points along the system centerlines can be calculated using the transport formula. Figure 9 provides an illustration of this process. The reconstructed centerlines can then be used to check for collisions in the binary occupancy map with inflated regions. It should be noted that the density of the reconstructed points should be relatively high for practical purposes, otherwise the inflated regions might clip the system centerlines without any reconstructed points being located inside them.

*E. Priority Queue Design*

As mentioned in a previous sub-section, the priority queue contains the details of all partial paths. For this proposed planning approach, the information included in each entry within the queue is listed in Table IV. Among these, action sequence, cost and terminal node states are necessary for the path-planning procedure to be carried out, with the action sequence providing the number of actions already applied to be used in cost function calculation, the cost informing on which node to expand first and terminal node states being used as initial conditions for node expansions. The remaining contents are included in the queue to allow for the easy visualization of the path-planning process. It should be remarked that even though the input history is available from the planning process and should theoretically allow the vehicle-trailer system to replicate the planned path, this is not a closed-loop system and hence is not robust against factors such as external disturbances. A feedback controller that will follow the desired path will be useful but is not treated here as it is outside the scope of the present paper.

TABLE IV
PRIORITY QUEUE CONTENTS

| Model Parameter | Value Choice |
|---|---|
| Action Sequence | Sequence of actions already applied in the current partial path |
| Cost | Cost value of the terminal node in the current partial path |
| Terminal Node States | Vehicle-trailer system states at the terminal node of current partial path |
| Expanded Branch Trajectory | Vehicle-trailer system trajectory of the most recent motion primitive in the current partial path |
| Overall Path Trajectory | Overall vehicle-trailer system trajectory of the current partial path |
| Input History | Vehicle-trailer system input history for the current partial path |

V. IMPLEMENTATION RESULTS

Simulation case studies are carried out to demonstrate the effectiveness of the proposed Hybrid A*-based path-planning approach. Table V lists the parameter value choices used in this case study. The parameter $T$ refers to the vehicle-trailer system simulation duration during the node expansion process where motion primitives are constructed, and its coupled usage with the value choice of constant trailer reverse speed $V_T$ guarantees that the motion primitive branches are comparable to each other using the cost function.

TABLE V
SIMULATION CASE STUDY VALUE CHOICES

| Model Parameter | Value Choice |
|---|---|
| $L$ | 2.896 [m] |
| $L_H$ | 1.159 [m] (passenger vehicle) |
| $L_T$ | 2.693 [m] |
| $Q$ | $\begin{bmatrix} 2 & 0 & 0 \\ 0 & 2 & 0 \\ 0 & 0 & 3 \end{bmatrix}$ |
| $K_A$ | 0.1 |
| $T$ | 1 [sec] |
| $V_T$ | -1 [m/sec] |
| $[\delta_{f,min}, \delta_{f,max}]$ | [-0.75, 0.75] [rad] = [-42.9718, 42.9718] [deg] |
| $[\delta_{T,lb2}, \delta_{T,ub2}]$ | [-0.5, 0.5] [rad] = [-28.6479, 28.6479] [deg] |

The binary occupancy map used in the case study is displayed in Figure 10. The goal of the path-planning procedure is to plan a kinematically feasible collision-free path from trailer start position to trailer goal position which is in an empty parking space with its neighboring parking spaces being occupied. Additionally, two inflated regions are added to facilitate the collision avoidance functionality of this proposed path-planning approach.

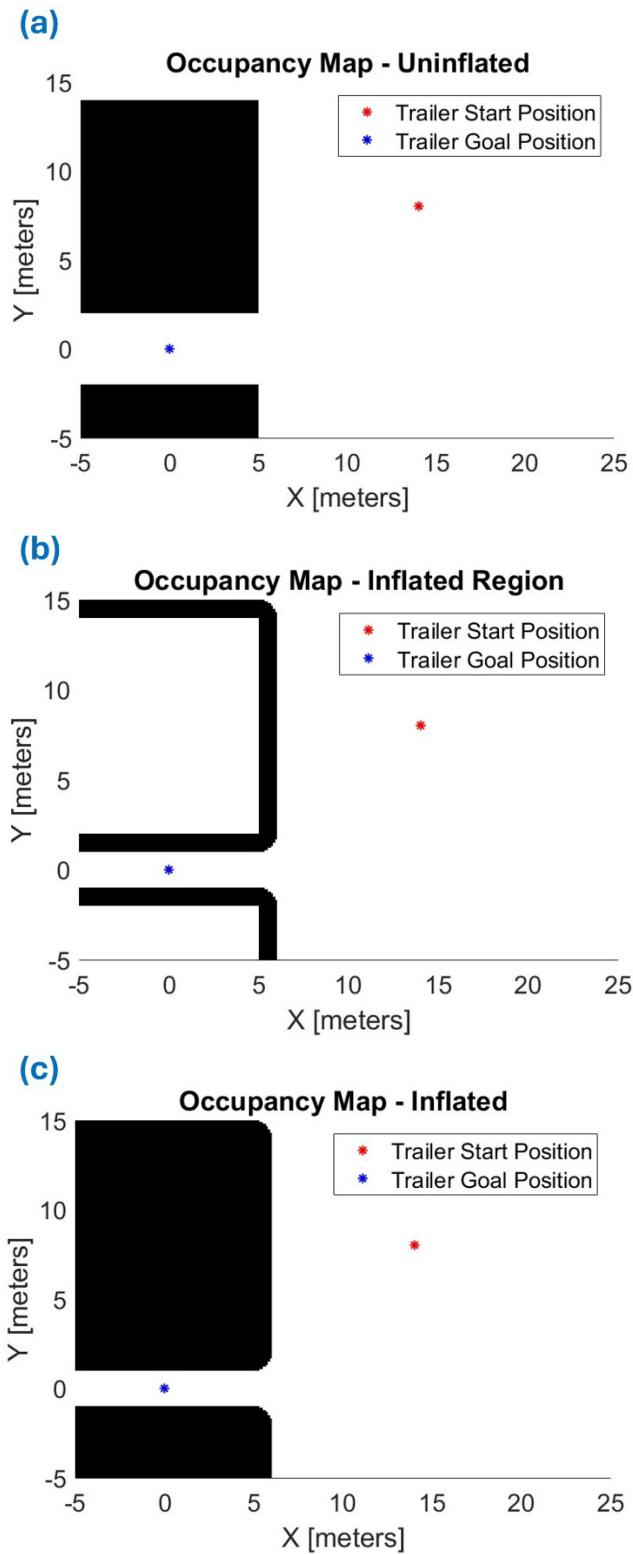

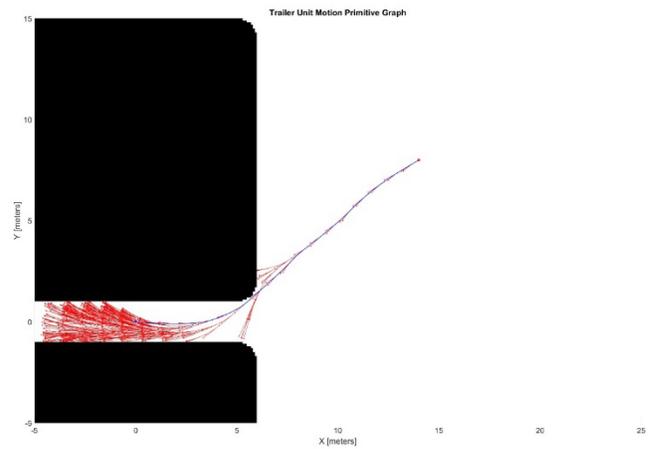

**Fig. 11.** Path-planning simulation result: motion primitive graph for the trailer unit.

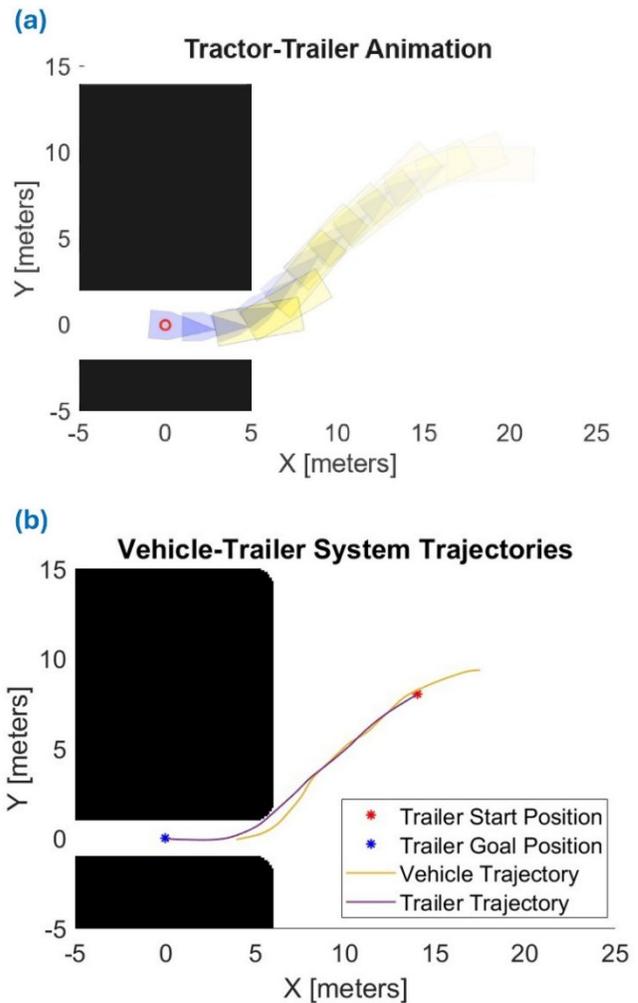

**Fig. 12.** Path-planning simulation Results: (a) Vehicle-trailer reverse motion; (b) Vehicle-trailer system trajectory.

**Fig. 10.** Binary occupancy map construction: (a) Original uninflated map; (b) Inflated map regions; (c) Modified inflated map.

The path-planning results are displayed in Figures 11-13. Figure 11 shows all the motion primitive branches generated for the trailer unit during the planning process. It can be observed that the motion primitives explored allow one sequence of branches to lead from trailer start position to a point very close to the trailer goal position, essentially forming a complete path that accomplishes the planning objective. This planned path is shown more clearly in Figure 12 with both vehicle unit and trailer unit trajectories displayed. It can be observed from Figure 12(b) that the system trajectories which describe the motion of the axle centers of both units clear the inflated binary occupancy map, which guarantees the exterior outlines of the units clear the uninflated occupancy map as demonstrated in Figure 12(a), proving the efficacy of the proposed collision avoidance design. It should be noted that the vehicle front axle steering input remains within the custom-defined limits, showcasing the effectiveness of the motion primitive design of the proposed path-planning algorithm. The method was further tested with many other starting positions which all performed effectively in achieving a collision free path ending close to the desired goal state.

## VI. Conclusion and Future Work

This paper proposed a Hybrid A*-based path-planning strategy to tackle the challenging problem of vehicle-trailer system reverse parking maneuver. A generic kinematic vehicle-trailer model with one trailer configuration was first derived. The inverse kinematics of the system was also derived to allow for the trailer to be regarded as a standalone unit where 'virtual' inputs at the trailer can later be propagated to the actual inputs at the tractor vehicle. A modified Hybrid A* algorithm was formulated to allow the planning of kinematically feasible and collision-free paths for the vehicle-trailer system in reverse motion. Simulation studies demonstrated the effectiveness of the proposed routine in its ability to guarantee kinematic feasibility and static obstacle avoidance. For future work, the path-planning approach proposed in this paper can be combined with path-tracking controllers such as nonlinear model predictive control (NMPC) systems to improve robustness of the parking operation. As an example of this integration, the path generated with the approach described in this paper can be used as a coarse initial trajectory in the NMPC routine that aims to mimic the initial path.

For future work, the path-planning approach proposed in this paper can be combined with path-tracking controllers such as nonlinear model predictive control (NMPC) systems [22], [23] or parameter space control systems [24], [25], [26], [27], [28] to improve robustness of the parking operation by introducing a feedback loop into the control system, rendering it a closed-loop operation. As an example of this integration, the path generated with the approach described in this paper can be used as a coarse initial trajectory in the NMPC routine that aims to mimic the initial path. Dynamic collision avoidance [29] and experiments with a real vehicle using their sensors [30], [31], [32], [33] to determine drivable areas and the parking spot will also be focus areas of future work. Simulated testing in realistic virtual environments can be used by creating such virtual settings [34]. Experimental testing of vehicle and trailer reverse parking can be performed safely and fast within immersed virtual environments using the Vehicle-in-Virtual-Environment Approach [35], [36], [37], [38]. Our future work will also explore the use of deep reinforcement learning based decision making [39], [40], [41], [42] for parking problems.

## Acknowledgment

The Ohio State University authors would like to thank HATCI (Hyundai America Technical Center, Inc.) for supporting this work. The Ohio State University authors thank the Automated Driving Lab for its support.